\documentclass[lettersize,journal]{IEEEtran}
\usepackage{amsmath,amsfonts}
\usepackage{algorithmic}
\usepackage{algorithm}
\usepackage{array}
\usepackage[caption=false,font=normalsize,labelfont=sf,textfont=sf]{subfig}
\usepackage[pagebackref,breaklinks,colorlinks=blue,citecolor=blue]{hyperref}
\usepackage{textcomp}
\usepackage{stfloats}
\usepackage{url}
\usepackage{verbatim}
\usepackage{graphicx}
\usepackage{cite}
\usepackage{amsmath}
\usepackage{amssymb}

\usepackage{makecell}
\usepackage{booktabs}
\usepackage{tabularray}
\UseTblrLibrary{diagbox}
\usepackage{tabularx}
\usepackage{diagbox}
\usepackage{xcolor}
\usepackage{xspace}
\usepackage{multirow}

\usepackage{pifont}

\newcommand{\cmark}{\ding{51}\xspace}%
\newcommand{\xmarkg}{\ding{55}\xspace}%

\usepackage{soul}
\sethlcolor{yellow}
\usepackage{xcolor}
\definecolor{yellow}{RGB}{0,0,0} 
\begin{document}

\title{MAT: Multi-Range Attention Transformer for\\ Efficient Image Super-Resolution}

\author{Chengxing Xie, Xiaoming Zhang, Linze Li, Yuqian Fu, Biao Gong, Tianrui Li,~\IEEEmembership{Senior Member,~IEEE,} Kai Zhang
\thanks{Manuscript received December 18, 2024; revised February 6, 2025; accepted March 17, 2025. This work was supported by the Science Fund Program for Distinguished Young Scholars (Overseas), and Sichuan Province Science and Technology Support Program (Grand Nos. 2024NSFTD0036, 2024ZHCG0166).  This article was recommended by Associate Editor Dr. Zhijun Lei. (\textit{Corresponding author: Kai Zhang}.)}        
\thanks{Chengxing Xie, Xiaoming Zhang, Linze Li, Tianrui Li are with School of Computing and Artificial intelligence, Southwest Jiaotong University, China. (email: \{zxc0074869, zxiaoming360, linze.longyue\}@gmail.com, trli@swjtu.edu.cn). Xiaoming Zhang is also with DAMO Academy, Alibaba Group, China.

Yuqian Fu is with INSAIT, Sofia University "St. Kliment Ohridski", Bulgaria. (e-mail: yuqian.fu@insait.ai).

Biao Gong is with Ant Group, China. (e-mail: a.biao.gong@gmail.com).

Kai Zhang is with the School of Intelligence Science
and Technology, Nanjing University, China
(e-mail: kaizhang@nju.edu.cn).

Chengxing Xie and Xiaoming Zhang contributed equally to this work.

Digital Object Identifier 10.1109/TCSVT.2025.3553135
}
}

\markboth{IEEE TRANSACTIONS ON CIRCUITS AND SYSTEMS FOR VIDEO TECHNOLOGY}%
{Shell \MakeLowercase{\textit{et al.}}: A Sample Article Using IEEEtran.cls for IEEE Journals}

\IEEEpubid{\begin{minipage}{\textwidth}\ \centering
		Copyright \copyright 2025 IEEE. Personal use of this material is permitted. \\
		However, permission to use this material for any other purposes must be obtained 
		from the IEEE by sending an email to pubs-permissions@ieee.org.
\end{minipage}}



\maketitle
\begin{abstract}
Image super-resolution (SR) has significantly advanced through the adoption of Transformer architectures. However, conventional techniques aimed at enlarging the self-attention window to capture broader contexts come with inherent drawbacks, especially the significantly increased computational demands. Moreover, the feature perception within a fixed-size window of existing models restricts the effective receptive field (ERF) and the intermediate feature diversity.  We demonstrate that a flexible integration of attention across diverse spatial extents can yield significant performance enhancements. In line with this insight, we introduce $\textbf{M}$ulti-Range $\textbf{A}$ttention $\textbf{T}$ransformer ($\textbf{MAT}$) for SR tasks. MAT leverages the computational advantages inherent in dilation operation, in conjunction with self-attention mechanism, to facilitate both multi-range attention (MA) and sparse multi-range attention (SMA), enabling efficient capture of both regional and sparse global features. Combined with local feature extraction, MAT adeptly capture dependencies across various spatial ranges, improving the diversity and efficacy of its feature representations. We also introduce the MSConvStar module, which augments the model's ability for multi-range representation learning. Comprehensive experiments show that our MAT exhibits superior performance to existing state-of-the-art SR models with remarkable efficiency ($\sim3.3\times$ faster than SRFormer-light). The codes are available at \href{https://github.com/stella-von/MAT}{https://github.com/stella-von/MAT}.
\end{abstract}



\section{Introduction}

\begin{figure}[t]
\centering
\includegraphics[width=0.99\linewidth]{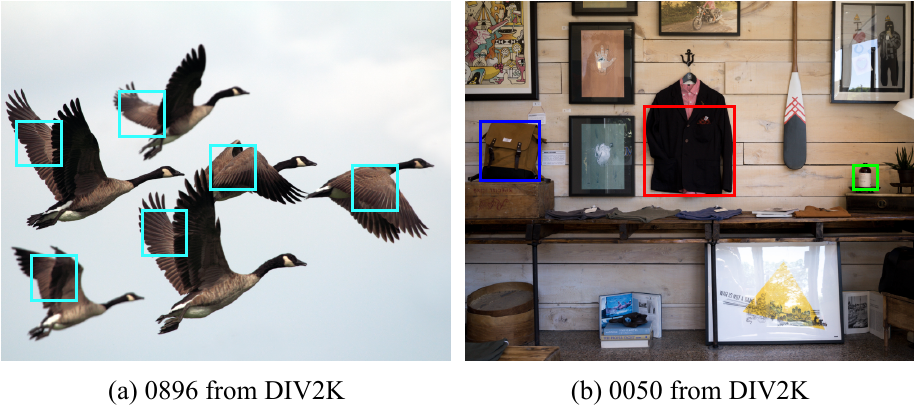}
\caption{(a) Illustration of image redundancy in natural images with self-similarity. Efficiently utilizing similar and repetitive structures and elements in natural images can aid in the reconstruction of image features. (b) In natural images, features can be observed to form a hierarchical structure across different spatial scales. The single and fixed-size WSA is insufficient to fully leverage such hierarchical features.}
\label{figs:image}
\vspace{-0.4cm}
\end{figure}

Single image super-resolution (SISR) is a long-standing problem in the low-level vision community that aims to restore a high-resolution (HR) image from its low-resolution (LR) counterpart. It is an ill-posed inverse problem with an infinite number of solutions since one LR image can be theoretically degraded from infinite HR images. To be well-posed,  it is essential to constrain the solution space by incorporating effective image priors. Two fundamental priors widely used for this purpose are image \textcolor[RGB]{65,105,225}{\textit{locality}} \cite{weinzaepfel2011reconstructing, zhang2023boosting} and \textcolor[RGB]{65,105,225}{\textit{redundancy}} \cite{shechtman2007matching, glasner2009super, 9467283}. The former indicates that image pixels exhibit strong correlations with their immediate neighbors while having weaker connections to distant pixels, while the latter suggests that similar patterns frequently recur at different locations within images\footnote{This phenomenon is also known as image self-similarity.}, as shown in Fig.~\ref{figs:image} (a). This manifests a conspicuous intuition that \textit{both local and non-local dependencies are beneficial for SISR tasks, but not all global dependencies}. These priors suggest that while both local and non-local dependencies contribute to SISR tasks, not all global dependencies are equally valuable. For example, when reconstructing one eye in a human face, the pixels within the target region are highly correlated, and the non-local information from the other eye provides useful reference features, while smooth areas like skin and background offer less meaningful information.
\begin{figure}[t!]
\centering
\includegraphics[width=0.99\linewidth]{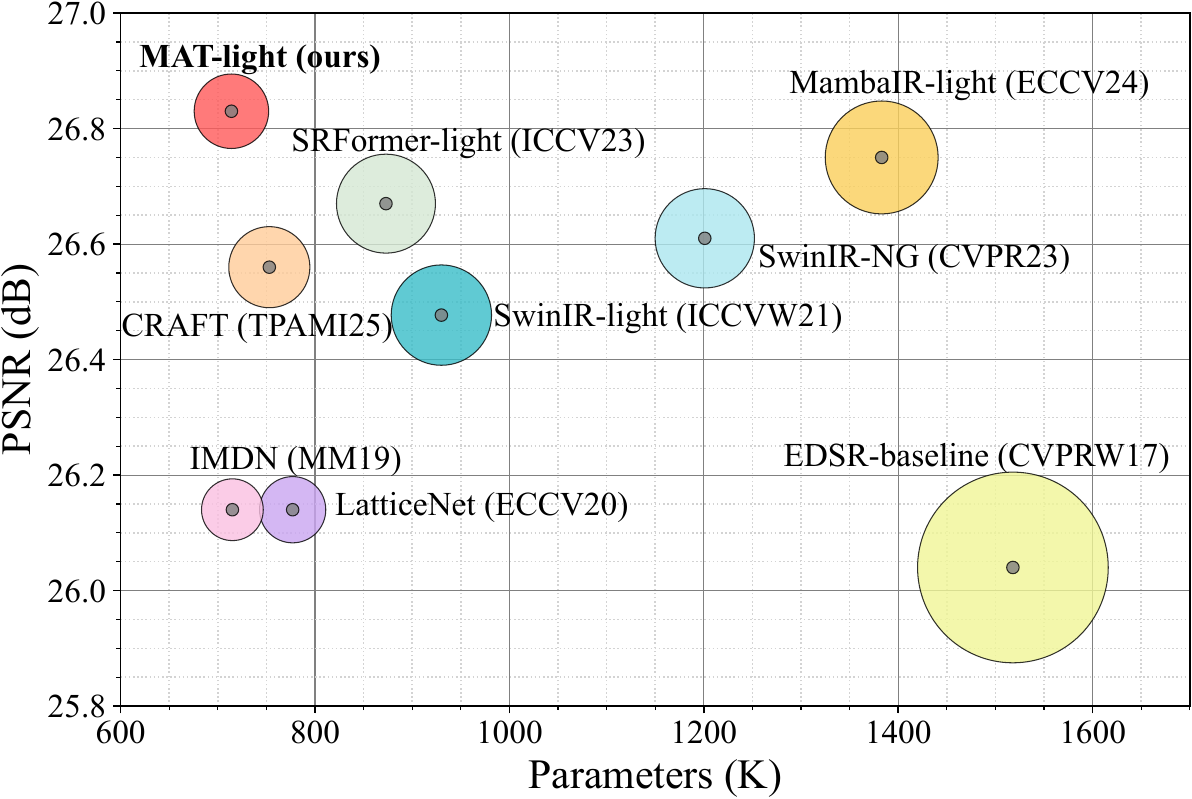}
\caption{Comparison of trade-offs between model performance and overheads on Urban100~\cite{huang2015single} for $\times 4$ SR. The area of each circle denotes the Multi-Adds of these models. Our MAT-light achieves optimal SR performance with fewer parameters and Multi-Adds.}
\label{figs:psnr}
\end{figure}

\IEEEpubidadjcol
In deep learning-based SISR methods, image locality is primarily preserved through convolutional operations, which extract local features within a fixed-size region (e.g., $k \times k$) using learnable kernels. This approach enables the regularization of the ill-posed inverse problem by learning locality priors from large-scale datasets. Various convolutional neural network (CNN) architectures have successfully implemented this principle, ranging from the pioneering SRCNN \cite{dong2015image} to more designs including residual networks \cite{kim2016accurate, ledig2017photo, lim2017enhanced}, dense connection models \cite{zhang2018residual, 9072448}, information distillation networks \cite{hui2019lightweight, li2022blueprint, xie2023large}, and attention-based approaches incorporating both channel \cite{zhang2018image} and spatial attention \cite{liu2020residualf, mei2021image}. CNN-based methods are inherently limited by their fixed-size convolutional kernels and local receptive fields, restricting their ability to capture broader contextual information and utilize global features effectively in HR image reconstruction.

Transformers~\cite{vaswani2017attention} employ self-attention mechanisms that excel at capturing long-range dependencies and extracting features across extensive spatial regions, making them particularly effective for SR tasks~\cite{gu2021interpreting}. Recent studies~\cite{liang2021swinir, chen2023activating, zhou2023srformer} have demonstrated the superiority of Transformer-based approaches over CNN-based methods in SR, particularly through enhanced window self-attention (WSA) mechanisms in Vision Transformers. However, expanding the window size to increase the effective receptive field (ERF)~\cite{ding2022scaling} leads to quadratic growth in computational complexity, creating significant efficiency challenges~\cite{zhou2023srformer}. Furthermore, while natural images inherently contain hierarchical features at various spatial scales, as illustrated in Fig.~\ref{figs:image} (b), most WSA-based models rely on homogeneous operators with fixed window sizes (e.g., $8 \times 8$ or $16 \times 16$). This limitation constrains their ability to effectively capture dependencies across different spatial ranges. Therefore, our primary goal is to develop an approach that enables efficient and flexible feature capture across multiple spatial ranges.

Building upon the principles of image \textcolor[RGB]{65,105,225}{\textit{locality}} and \textcolor[RGB]{65,105,225}{\textit{redundancy}}, we recognize that effectively processing features at different spatial ranges while \textbf{selectively} utilizing non-local information is crucial for SR tasks. To leverage this insight, we develop an approach that exploits hierarchical features at local, regional, and global levels to enhance the model's multi-range representation learning capacity\textcolor{yellow}{~\cite{li2023efficient}}. First, we construct a Local Aggregation Block (LAB)  that combines convolution and channel attention~\cite{zhang2018image} for efficient local feature aggregation. Inspired by the success of commonly-used dilated convolutions~\cite{chen2017rethinking, wang2018understanding}, we incorporate dilation operations into the attention mechanism to expand the perceptual scope to both regional and global levels without increasing computational complexity.

We first replace standard convolution computation with an attention mechanism to achieve regional attention. Then, by introducing holes in the attention region, similar to dilated convolutions, we implement efficient sparse global attention. To overcome the limitations of homogeneous operators in multi-scale feature capture, we extend these concepts into multi-range attention (MA) and sparse multi-range attention (SMA). Both MA and SMA operate across multiple regional ranges, achieving a larger effective receptive field than fixed-size attention windows. Additionally, we introduce the MSConvStar module as a replacement for the traditional feed-forward network (FFN), enhancing image token interactions through the integration of multi-scale convolution (MSConv) with star operation~\cite{Ma_2024_CVPR}.

Based on the aforementioned designs, we present the Multi-Range Attention Transformer (MAT), a novel architecture for image SR that effectively captures features across diverse spatial ranges to enhance detail reconstruction. The lightweight variant, MAT-light, achieves state-of-the-art performance with remarkable efficiency when trained solely on the DIV2K~\cite{agustsson2017ntire} dataset,  achieving state-of-the-art performance with lower computational complexity (e.g., \textbf{26.83dB}@Urban100 $\times 4$ with only \textbf{714K} parameters), as shown in Fig.~\ref{figs:psnr}. Similarly, in classical image SR tasks, MAT demonstrates superior performance while maintaining lower computational complexity. The key contributions of this work are:

\noindent (1) We introduce MA and SMA, two complementary mechanisms that enable flexible capture of multi-range regional characteristics and sparse global attributes, advancing multi-range representation learning in SR tasks.

\noindent (2) We develop the MSConvStar module, an efficient alternative to conventional FFNs, designed specifically to enhance multi-range feature capture through the integration of hierarchical feature processing.


\noindent (3) We propose \textbf{MAT}, a novel and computationally efficient SR architecture that consistently outperforms the current state of the art while requiring fewer computational resources, as validated through extensive experimentation.


\section{Related Works}

\subsection{CNN-based Image SR}

CNNs have dominated SR approaches since the introduction of SRCNN~\cite{dong2015image}. VDSR~\cite{kim2016accurate} employs a deeper network architecture, integrating residual learning to enhance performance metrics. \textcolor{yellow}{EDSR~\cite{lim2017enhanced} enhances the generalization ability of the model by removing the BatchNorm~\cite{pmlr-v37-ioffe15} layers, which were found to impede SR performance.} RDN~\cite{zhang2018residual} implements dense connections for better hierarchical feature utilization, while RCAN~\cite{zhang2018image} introduces channel attention to dynamically adjust feature importance. IGNN~\cite{zhou2020cross}, NLSN~\cite{mei2021image}, and FPAN~\cite{10056962} incorporate non-local attention~\cite{wang2018non} to capture global image characteristics. For resource-constrained applications, several approaches~\cite{hui2019lightweight, li2022blueprint} employ information distillation to create lightweight models. However, despite various enhancements including attention mechanisms~\cite{zhang2018image, mei2021image}, large kernel convolutions~\cite{xie2023large}, and partial channel shifting~\cite{zhang2023boosting}, CNN-based SR networks remain limited in their information perception range, struggling to model long-range dependencies~\cite{chen2023activating}.

\begin{figure*}[!ht]
\centering
\includegraphics[width=0.98\linewidth]{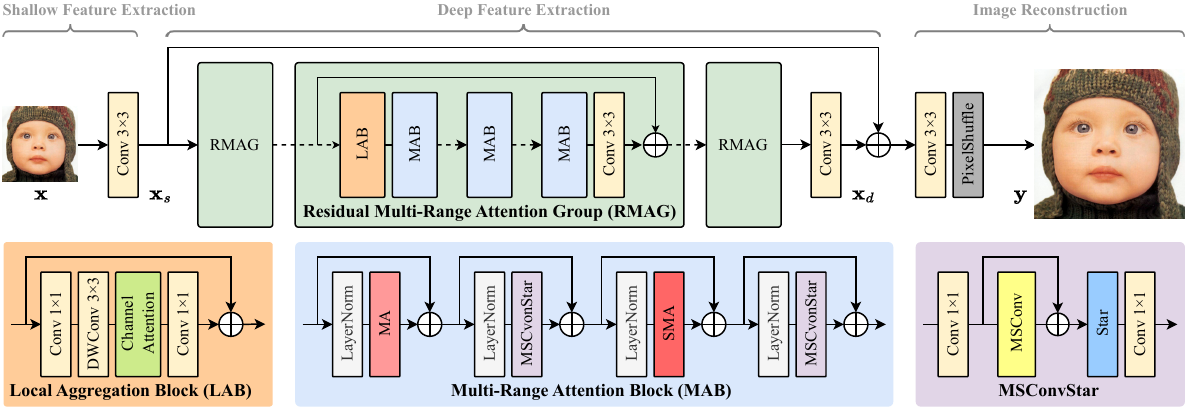}
\caption{The overall architecture of \textbf{M}ulti-Range \textbf{A}ttention \textbf{T}ransformer (\textbf{MAT}).}
\label{figs:mat}
\vspace{-0.3cm}
\end{figure*}

\subsection{Transformer-based Image SR}
\textcolor{yellow}{Vision Transformers have demonstrated remarkable success across diverse visual tasks~\cite{dosovitskiy2020image, liu2021swin, xie2021segformer, carion2020end, zamir2022restormer, 10050878, 10138555, li2023lightweight, wang2023hybrid, lu2023cross}.} Their superior ability to capture long-range dependencies and extract features from extensive regions makes them particularly effective for SR tasks. SwinIR~\cite{liang2021swinir} successfully adapts Swin-Transformer~\cite{liu2021swin} architecture for image restoration, while Omni-SR~\cite{wang2023omni} enhances performance by simultaneously modeling spatial and channel interactions. \textcolor{yellow}{While ART~\cite{zhang2023accurate} combines dense and sparse attention mechanisms to expand the receptive field, it suffers from implementation inefficiencies. HAT~\cite{chen2023activating} and SRFormer~\cite{zhou2023srformer} improve performance through enlarged self-attention windows.} However, WSA-based methods face efficiency challenges as window size increases, resulting in quadratic computational growth and higher memory requirements. Additionally, fixed-size window partitioning limits flexible multi-scale information integration. Our work addresses these limitations by incorporating dilation operations into self-attention mechanisms, enabling flexible perception range expansion without additional computational overhead.

\subsection{Hierarchical Feature Representation}
\textcolor{yellow}{
Hierarchical feature representation has emerged as a fundamental component in diverse visual tasks~\cite{hui2019lightweight,li2023transformer,pan2023slide,jiao2023dilateformer,li2023efficient}. Recent approaches have explored various strategies to leverage this concept: DMFN~\cite{li2023transformer} achieves multi-scale feature perception through coordinated upsampling and downsampling operations, while Slide-Transformer~\cite{pan2023slide} and DilateFormer~\cite{jiao2023dilateformer} enhance the receptive field of $3\times3$ attention windows using deformed shifting and dilation operations, respectively. Although these methods demonstrate progress, their effective receptive fields remain limited, particularly when processing high-resolution images. While GRL~\cite{li2023efficient} addresses this limitation by introducing anchored stripe self-attention and implementing a global-regional-local framework, its architectural complexity impedes practical deployment. In contrast, our approach achieves superior hierarchical feature representation while maintaining computational efficiency.}


\section{Methodology}
\subsection{Overall Architecture of MAT}
Following established approaches~\cite{liang2021swinir, chen2023activating}, our proposed MAT architecture, illustrated in Fig.~\ref{figs:mat}, comprises three primary modules: shallow feature extraction, deep feature extraction, and image reconstruction.  Given an LR input $\mathbf{x} \in \mathbb{R}^{H \times W \times 3}$, shallow features $\mathbf{x}_s \in \mathbb{R}^{H \times W \times C}$ are first extracted using a $3 \times 3$ convolutional layer:
\begin{equation}
\mathbf{x}_s = H_{SF}(\mathbf{x}),
\end{equation} where $C$ denotes the channel dimension and $H_{SF}(\cdot)$ represents the convolution operation. The shallow features ${\mathbf{x}_s}$ then is fed into the deep feature extraction through a series of residual multi-range attention groups (RMAG) followed by a $3\times3$ convolution, denoted as $H_{DF}$, to produce deep features ${\mathbf{x}_d \in \mathbb{R}^{H \times W \times C}}$:
\begin{equation}
\mathbf{x}_d = H_{DF}(\mathbf{x_s}).
\end{equation} Each RMAG integrates a local aggregation block (LAB), multiple multi-range attention blocks (MAB), a $3\times3$ convolution, and a residual connection, as depicted in Fig.~\ref{figs:mat}. The shallow and deep features are combined to enhance convergence, followed by the reconstruction module $H_{RC}$ (comprising convolution and PixelShuffle~\cite{shi2016real}) to generate the final HR image $\mathbf{y} \in \mathbb{R}^{H \times W \times 3}$:
\begin{equation}
\mathbf{y} = H_{RC}(\mathbf{x_s}+\mathbf{x_d}).
\end{equation}
Notably, this architectural framework aligns with established approaches~\cite{zhang2018image,liang2021swinir,zhou2023srformer,wang2023omni}, ensuring fair comparison with off-the-shelf methods. The model is optimized with $\mathcal{L}_1$ loss between the reconstructed image and the ground-truth HR image:
\begin{equation}
L(\theta) = \arg\min_{\theta} \frac{1}{N} \sum_{i=1}^{N} \|H_{MAT}(I_{LR}^i;\textcolor{yellow}{\theta}) - I_{HR}^i \|_1,
\end{equation} where $H_{MAT}(\cdot)$ denotes MAT model, $\theta$ denotes the learnable parameters, and $N$ is the number of LR-HR training pairs.

\subsection{Attention Mechanisms}
\noindent \textbf{Self-Attention.}
Self-attention (SA)~\cite{vaswani2017attention} transforms input sequences through weighted aggregation of value vectors, where weights are determined by query-key interactions. Given a query $\mathbf{Q}$, and corresponding key-value pairs $\mathbf{K}$ and $\mathbf{V}$, SA computes scaled dot-product attention, normalized through a softmax function, to generate attention weights for value aggregation. This process can be formally expressed as: 
\begin{equation}
\operatorname{SA}(\mathbf{Q},\mathbf{K},\mathbf{V})=\operatorname{Softmax}\left(\mathbf{QK}^{T}/{\sqrt{d}+B}\right)\mathbf{V},
\end{equation}
where $d$ denotes the embedding dimension and $B$ represents a learnable relative positional bias. While SA effectively captures global features and enhances texture details, its direct application to SR tasks presents two significant challenges: potential over-weighting of noise information and substantial computational overhead~\cite{mei2021image, xia2022efficient}.

\begin{figure*}[!ht]
\centering
\includegraphics[width=0.98\linewidth]{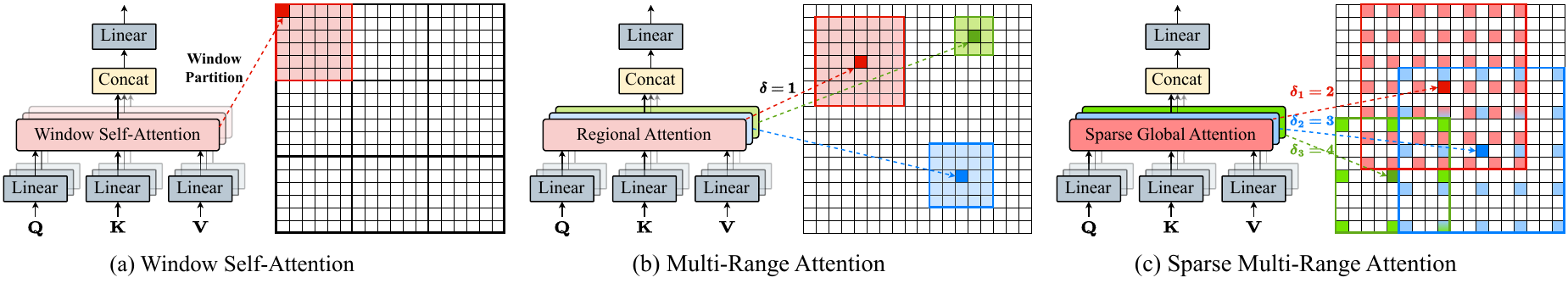}
\caption{Illustration of the window self-attention (WSA), multi-range attention (MA) and sparse multi-range attention (SMA). MA and SMA set different range sizes for different attention heads, enabling the multi-range representation learning.}
\label{figs:attention}
\end{figure*}

\begin{figure*}[!ht]
\centering
\includegraphics[width=0.98\linewidth]{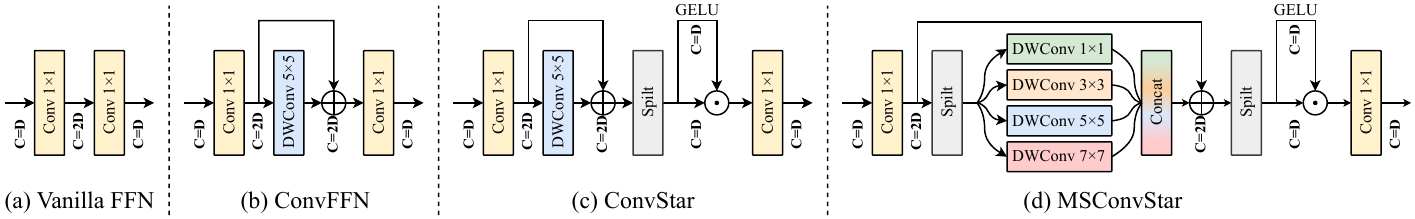}
\caption{Illustration of FFN~\cite{liang2021swinir}, ConvFFN~\cite{zhou2023srformer}, ConvStar and  MSConvStar. $\odot$: element-wise multiplication (star).}
\label{figs:msconvstar}
\end{figure*}

\noindent\textbf{Multi-Range Attention.}
While WSA has been widely adopted~\cite{dosovitskiy2020image, liu2021swin} to balance computational efficiency and performance (Fig.~\ref{figs:attention}~\textcolor{red}{(a)}), its effectiveness is limited by rigid window partitioning and computational constraints, which restrict flexible window sizing and broader spatial feature perception. To address these limitations, we propose restricting attention computation to specific neighborhoods, analogous to convolution operations, implementing a regional attention (RA) mechanism through sliding calculations. For RA with range size $k$, the key-value set for pixel $p_{i,j}$ at position $(i,j)$ is confined to a $k \times k$ region, denoted as $\rho^k_{i,j}$. Given a feature map of dimensions $H \times W$, the RA computation for this pixel is expressed as:
\begin{equation}
\operatorname{RA}_k(p_{i,j})=\operatorname{Softmax}\left(\mathbf{Q}_{i,j}\mathbf{K}^{T}_{\rho^k_{i,j}}/{\sqrt{d}}+B\right)\mathbf{V}_{\rho^k_{i,j}},
\end{equation} where $1 \le i \le W$ and $1 \le j \le H$. Complete RA is achieved by applying this computation across all feature pixels. While various implementations of RA exist, including SASA~\cite{ramachandran2019stand}, natten~\cite{hassani2023neighborhood}, and HaloNet~\cite{vaswani2021scaling}, we adopt the efficient natten~\cite{hassani2023neighborhood} approach. To overcome the limitations of homogeneous operators in multi-scale feature capture, we extend RA to multi-range attention (MA) as shown in (Fig.~\ref{figs:attention}~\textcolor{red}{(b)}):
\begin{equation}
\operatorname{MA}(p_{i,j})=H_F\left(\operatorname{Concat}\left(\operatorname{RA}_{k_1},\ldots,\operatorname{RA}_{k_n}\right)\right),
\end{equation}
where $k_1,\ldots,k_n$ represent $n$ different regional ranges, and $H_F$ denotes the feature fusion module. By simultaneously processing information from multiple spatial ranges, MA effectively addresses the constraints of fixed-size window partitioning in WSA.

\noindent\textbf{Sparse Multi-Range Attention.}
MA can be generalized to sparse multi-range attention (SMA), akin to dilated convolutions, as shown in Fig~\ref{figs:attention}. Dilating the region prompts the model to capture an extended array of associations for the target pixel. We expand RA through dilation operation to achieve sparse global attention (SGA). Specifically, for SGA with a range size of $k$ and a dilation rate of $\delta$, the key-value set corresponding to the $(i,j)$-th pixel $p_{i,j}$ is limited to a sparse neighborhood of size $k_d \times k_d$, denoted as $\rho^{k,\delta}_{i,j}$, where $k_d=k+(k-1)\cdot(\delta-1)$. The SGA of the pixel $p_{i,j}$ can be defined as:
\begin{equation}
\operatorname{SGA}^{\delta}_k(p_{i,j})=\operatorname{Softmax}\left({\mathbf{Q}_{i,j}\mathbf{K}^{T}_{\rho^{k,\delta}_{i,j}}}/{\sqrt{d}}+B\right)\mathbf{V}_{\rho^{k,\delta}_{i,j}}.
\end{equation}
The keys and values corresponding to the $(i,j)$-th pixel $p_{i,j}$ will be selected from the following set \textcolor{yellow}{$(i^{'}, j^{'})$} for self-attention computation:
\begin{equation}
\textcolor{yellow}{
\left\{(i^{'}, j^{'}) \;\middle|\; i^{'} = i + x \times \delta, j^{'} = j + y \times \delta \right\},}
\end{equation}
where $-\frac{k}{2} \leq x, y \leq \frac{k}{2}$. Subsequently, by aggregating SGAs from multiple ranges, we achieve SMA:
\begin{equation}
\operatorname{SMA}(p_{i,j})=H_F\left(\operatorname{Concat}\left(\operatorname{SGA}_{k_1}^{\delta_1},\ldots,\operatorname{SGA}_{k_n}^{\delta_n}\right)\right),
\end{equation}
where $\delta_1, \ldots, \delta_n$ represent $n$ types of dilation rates. The sparse area size can be flexibly enlarge by adjusting dilation rates, all while without introducing computational overhead.

\subsection{Multi-Range Representation Learning}
\noindent \textbf{MSConvStar.}
The conventional feed-forward network (FFN) in Transformer architectures~\cite{vaswani2017attention} utilizes two linear projections with an activation function, essentially implementing a basic multi-layer perceptron (MLP) for image token interaction. However, this structure proves inadequate for modeling complex, hierarchical spatial relationships, limiting SR model performance. While recent approaches~\cite{zamir2022restormer, zhou2023srformer} incorporate convolutions into MLPs to enhance spatial feature learning, they operate at a single scale. To address these limitations, we propose the multi-scale convolution star (MSConvStar) module, which enhances token representation through the integration of multi-scale convolution (MSConv) and star operations~\cite{Ma_2024_CVPR}, as illustrated in Fig.~\ref{figs:msconvstar}. This module combines parallel depth-wise convolutions at different scales with residual connections, working synergistically with multi-range attention to enhance spatial relationships.  The star operation also enables nonlinear feature space transformation without increasing network width, thereby improving the model's expressive capacity.

\noindent\textbf{Multi-Range Dependencies.}
Most SR models~\cite{lim2017enhanced, liang2021swinir, zhou2023srformer, 10852524} typically overlook the inherent hierarchical features in natural images, relying instead on homogeneous operators for structural modeling. Our MAT addresses this limitation by explicitly modeling three distinct types of dependencies across different spatial scales through specialized components: LAB, MA, and SMA (Fig.~\ref{figs:mat}). LAB captures local neighborhood dependencies through depth-wise convolutions and channel attention~\cite{zhang2018image}, while MA and SMA model regional attributes and sparse global information, respectively. Recognizing that long-range dependencies are relatively weaker, we incorporate MA and SMA within a Transformer framework~\cite{vaswani2017attention}, replacing the traditional MLP with our MSConvStar to create the MAB module for enhanced feature learning. This comprehensive approach enables more effective spatial structure modeling, enriches intermediate feature diversity, and ultimately yields more precise detail reconstruction.


\section{Experiments}\label{sec:experiments}

\subsection{Experimental Settings}
\noindent \textbf{Datasets and Evaluation.}
Follow previous works~\cite{liang2021swinir, zhou2023srformer}, we train two versions of MAT: lightweight and classical. For lightweight image SR, we utilize DIV2K~\cite{agustsson2017ntire} dataset to train our MAT-light. For classical version, the DF2K (DIV2K~\cite{agustsson2017ntire} + Flicker2K~\cite{lim2017enhanced}) is employed. Our evaluation of the models is performed on five commonly used benchmark datasets, including Set5~\cite{bevilacqua2012low}, Set14~\cite{zeyde2012single}, B100~\cite{martin2001database}, Urban100~\cite{huang2015single}, and Manga109~\cite{matsui2017sketch}. The PSNR (dB) and SSIM scores, calculated on the luminance (Y) channel, are used to evaluate the performance of the model.

\noindent \textbf{Implementation Details.}
For the lightweight image SR, we set the number of RMAG, MAB and channel to 4, 2, and 60, respectively. The range sizes of MA and SMA are both set to $7 \times 7$, $9 \times 9$, and $11 \times 11$. The dilation rates of SMA are set to the floor division of the input patch size ($64 \times 64$) by the range sizes, i.e., 9, 7, and 5. The total number of attention heads is set to 6, with two heads allocated to each range size. For the classical image SR, the the number of RMAG, MAB and channel increase to 6, 3, and 156, respectively. The range sizes increase to $13 \times 13$, $15 \times 15$ and $17 \times 17$. The training patch size is set to $64 \times 64$. For data augmentation, we randomly rotate and horizontally flip the input patches. We employ Adam~\cite{kingma2014adam} optimizer with $\beta_1=0.9$ and ${\beta}_2=0.99$ to train the model, with a total of 500k iterations. The initial learning rate is set at $2 \times 10 ^{-4}$ and is halved at [250k, 400k, 450k, 475k].

\subsection{Ablation Study}
We conduct comprehensive ablation experiments using lightweight SR models trained from scratch on DIV2K ($\times2$) and evaluated on Urban100 for $\times2$ SR. To ensure fair comparison, all models maintain consistent training protocols and hyperparameters. Additionally, we provide the Multi-Adds metric for each model, which is computed based on upscaling a single image input to a resolution of $1280 \times 720$.

\begin{table}[t!]
\centering
\setlength\tabcolsep{4pt}
\caption{Ablation study on each component.}
\label{table:component}
\scalebox{0.97}{
\begin{tabular}{ccc|ccc}
\bottomrule
MD & MSConvStar & \textcolor{yellow}{MR} & \makecell{Params.\\ (K)} & \makecell{Multi-Adds \\ (G)} & \makecell{Urban100 \\ PSNR (dB) / SSIM} \\ \hline
\xmarkg & \xmarkg & \xmarkg & 670 & 185.2 & 32.59 / 0.9329 \\
\cmark & \xmarkg & \xmarkg & 709 & 192.4 & 32.93 / 0.9357 \\
\cmark & \cmark & \xmarkg & 693 & 188.4 & 33.11 / 0.9374 \\
\cmark & \cmark & \cmark & 694 & 189.6 & \textbf{33.22} / \textbf{0.9381} \\ \toprule
\end{tabular}}
\vspace{-0.2cm}
\end{table}

\noindent \textbf{Effectiveness of Each Component.}
We systematically evaluate the contribution of each proposed component through detailed ablation experiments, as presented in Table~\ref{table:component}. \textbf{\textit{Firstly, baseline configuration.}} Our baseline model adapts SwinIR-light by replacing its Transformer blocks with regional attention ($9 \times 9$ range size, comparable to SwinIR-light's window size) and reducing the number of Transformer blocks to four per layer. \textbf{\textit{\textcolor{yellow}{Secondly, multi-range dependencies (MD).}}} Integration of LAB and sparse global attention enables feature capture across diverse spatial ranges, significantly enhancing model performance. \textbf{\textit{Thirdly, MSConvStar integration.}} Replacing the conventional MLP with MSConvStar improves feature representation capability while reducing both parameter count and computational overhead. \textbf{\textit{ \textcolor{yellow}{Finally, multi-range (MR) strategy.}}} Extending single-region attention to multi-range attention expands the model's receptive field, yielding our final MAT-light architecture, improved performance with minimal computational cost increase. These sequential improvements demonstrate the effectiveness of each component, as further validated through reverse-order ablation experiments.

\begin{figure*}[!t]
\centering
\vspace{0cm}
\setlength{\abovecaptionskip}{0.0cm}
\includegraphics[width=0.93\linewidth]{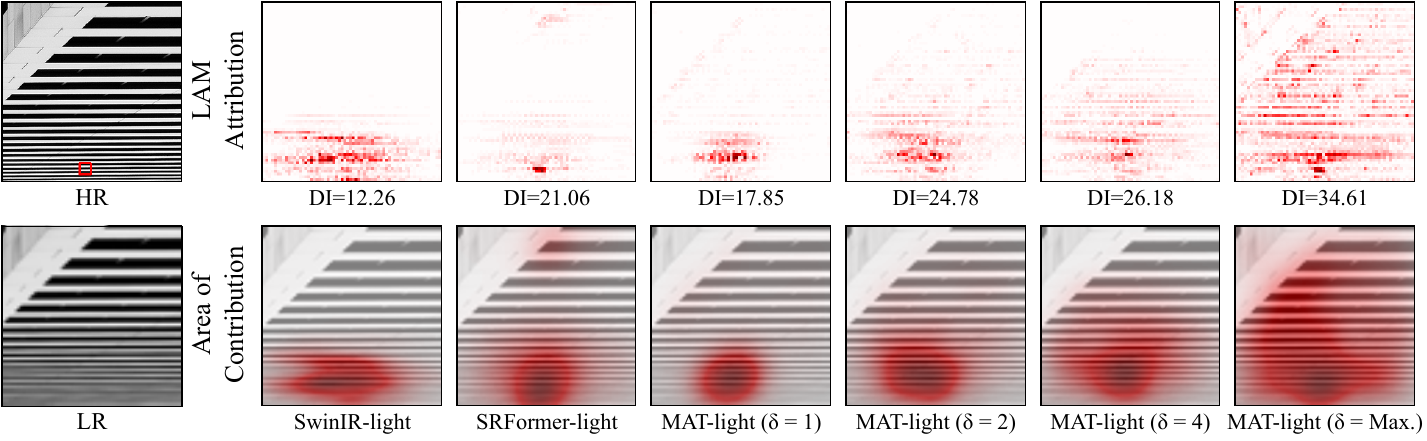}
\caption{LAM~\cite{gu2021interpreting} results of SwinIR~\cite{liang2021swinir}, SRFormer~\cite{zhou2023srformer} and our MAT with various dilation rates. The experiments is conducted using lightweight models. The LAM attribution indicates the importance of each pixel in LR input with respect to the SR result of the patch marked with a red box. The diffusion index (DI)~\cite{gu2021interpreting} reflects the range of involved pixels, and a higher DI indicates a broader range of pixels available to the model. $\delta$ denotes the dilation rate of MAT-light. The results indicate that, SRFormer-light, due to the use of larger windows (i.e. $16 \times 16$), can leverage more information compared to SwinIR-light. Additionally, MAT-light can use the most pixels for reconstruction by increasing the dilation rate without introducing additional computational burden.}
\label{figs:lam}
\vspace{-0.5cm}
\end{figure*}

\noindent \textbf{Effects of Dilation Rate.}
The dilation strategy in MAT enables broader information capture, enhancing overall performance. We investigate the optimal dilation rate $\delta$ through systematic experimentation, with results presented in Table~\ref{table:dilation}. Here, $\delta=1$ represents standard MA, while $\delta=$ Maximum indicates SMA with dilation rate set to the floor division of the input feature map size by the range size. Our experiments reveal a positive correlation between performance and dilation rate, with optimal results achieved at $\delta=$ Maximum. \textcolor{yellow}{However, further increasing the dilation rate through padding leads to significant performance degradation, likely due to the introduction of irrelevant elements that compromise attention weight learning.} Visual analysis using LAM~\cite{gu2021interpreting} (Fig.~\ref{figs:lam}) demonstrates that larger dilation rates expand the model's perceptive field. Notably, MAT-light's LAM attribution spans nearly the entire image, while competing models exhibit more restricted ranges of influence.

\begin{table}[!t]
\setlength\tabcolsep{2pt}
\caption{Effects of dilation rate $\delta$. Note $\delta=$ Maximum indicates that the dilation rate for SMA is set to the floor division of the input feature map size ($64 \times 64$) by the range sizes (7, 9, 11).}
\label{table:dilation}
\centering
\begin{tabular}{l|ccccc}
\bottomrule
\textcolor{yellow}{\makecell{Dilation Rate \\ $\{\delta_1, \delta_2, \delta_3\}$}} & $\{1,1,1\}$ & $\{2,2,2\}$ & $\{4,4,4\}$ & \makecell{$\{5,7,9\}$ \\ (Maximum)} & \textcolor{yellow}{\makecell{$\{6,8,10\}$ \\ (Exceeding)}} \\ \hline
PSNR (dB) & 32.94  & 33.11 & 33.16 & \textbf{33.22} & 29.60 \\
SSIM & 0.9358 & 0.9373 & 0.9378 & \textbf{0.9381} & 0.8743 \\
\toprule
\end{tabular}
\vspace{-0.4cm}
\end{table}

\begin{figure}[!t]
\centering
\setlength{\abovecaptionskip}{0.0cm}
\includegraphics[width=0.99\linewidth]{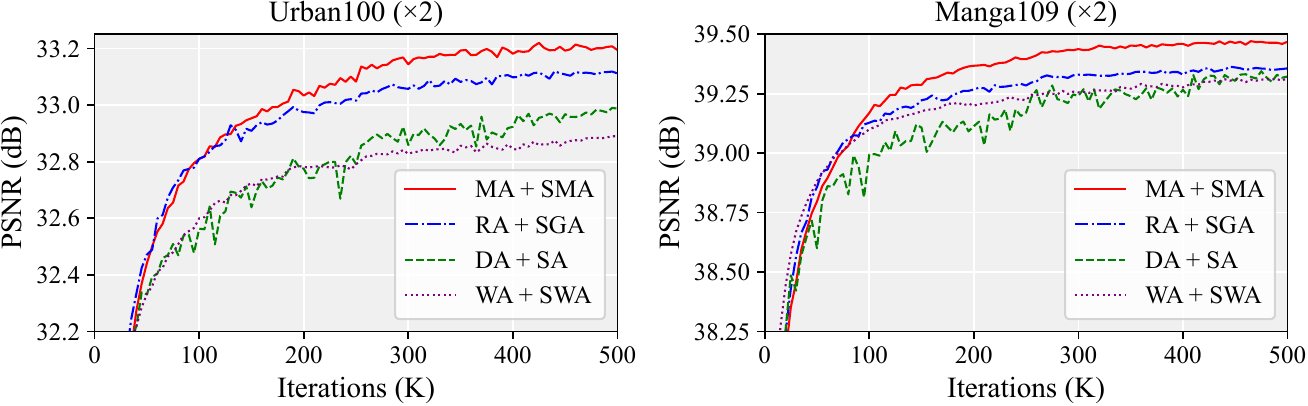}
\caption{PSNR (dB) comparison of different attention mechanisms on Urban100 and Manga109 datasets.}
\vspace{-0.3cm}
\label{figs:plot_attention}
\end{figure}

\noindent \textbf{Effectiveness of Multi-Range Attention.}
We evaluate our multi-range attention (MA) mechanism against representative attention mechanisms: window attention (WA)~\cite{liang2021swinir}, sparse attention (SA)~\cite{zhang2023accurate}, and regional attention (RA). As demonstrated in Table~\ref{table:attention} and Fig.~\ref{figs:plot_attention}, ART's combination of dense attention (DA) and sparse attention (SA) outperforms SwinIR's window attention (WA) and shifted window attention (SWA), validating the effectiveness of sparse operations. \textcolor{yellow}{However, ART's~\cite{zhang2023accurate} window partition-based sparse operation implementation limits token interval flexibility and incurs substantial computational overhead. Regional attention (RA) and sparse global attention (SGA) address these window partitioning limitations in DA and SA, yielding improved model performance.} The extension to MA and SMA further enhances this improvement, achieving optimal performance metrics.

\begin{table}[!t]
\centering
\setlength\tabcolsep{6pt}
\caption{Ablation study on attention mechanisms.}
\label{table:attention}
\scalebox{1}{
\begin{tabular}{l|ccc}
\bottomrule
Attention Type & \makecell{Params.\\ (K)} & \makecell{Multi-Adds \\ (G)} & \makecell{Urban100 \\ PSNR (dB) / SSIM}  \\ \hline
WA + SWA & 687  & 179.2 & 32.89 / 0.9348 \\
DA + SA & 667  & $>$280 & 32.99 / 0.9358 \\
RA + SGA & 693  & 188.4 & 33.11 / 0.9374 \\
MA + SMA & 694  & 189.6 & \textbf{33.22} / \textbf{0.9381} \\
\toprule
\end{tabular}}
\vspace{-0.3cm}
\end{table}

\begin{figure}[!t]
\centering
\setlength{\abovecaptionskip}{0.0cm}
\includegraphics[width=0.99\linewidth]{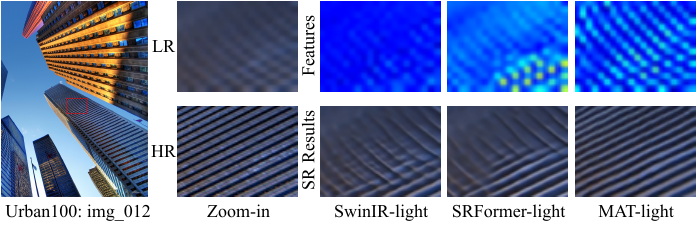}
\caption{Visual comparison of features and reconstruction results on $\times 4$ SR. MAT-light captures \textbf{holistic} structural patterns more effectively, highlighting its retention of crucial image details. }
\label{figs:feature}
\vspace{-0.5cm}
\end{figure}

To further validate our approach, we visualize pre-upsampling feature maps across models with different attention strategies (Fig.~\ref{figs:feature}). MAT-light demonstrates superior texture preservation and detail retention in feature representations compared to alternative models, leading to sharper reconstruction results. These visualizations provide additional evidence that multi-range attention significantly enhances feature representation capabilities.

\noindent \textbf{Effectiveness of MSConvStar.}
We evaluate MSConvStar through comparative analysis of four distinct configurations illustrated in Fig.~\ref{figs:msconvstar}. Results in Table~\ref{table:msconvstar} demonstrate that integrating depth-wise convolutions into MLP enhances performance, underscoring the significance of spatial information modeling. The star operation further improves performance while reducing model complexity through enhanced non-linear expression. Our MSConvStar extends this framework to multi-scale convolutions, enabling richer feature capture across varied scales.

\begin{figure*}[!t]
\centering
\setlength{\abovecaptionskip}{0.1cm}
\includegraphics[width=0.94\textwidth]{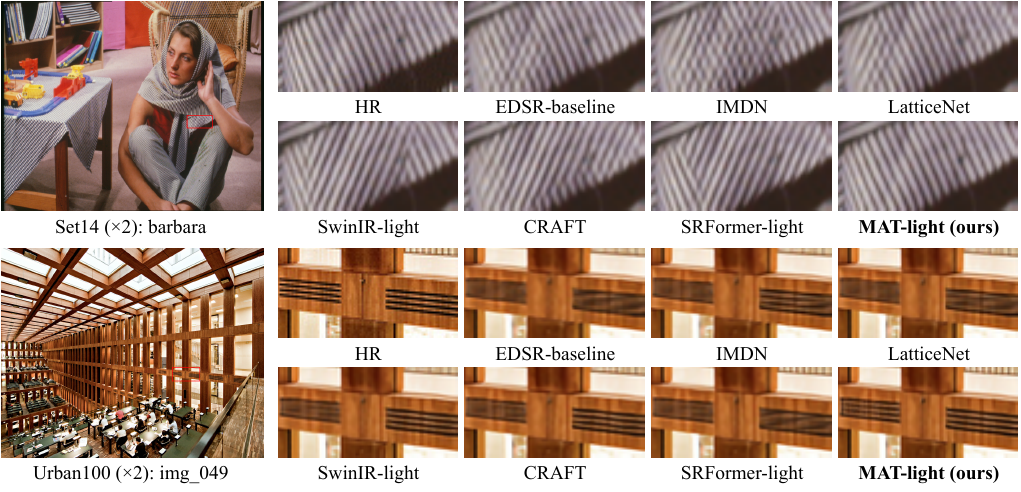}
\caption{Visual comparison on $\times 2$ \textbf{lightweight image SR}. The patches for comparison are marked with red boxes in the original images.}
\label{figs:visual1}
\vspace{-0.4cm}
\end{figure*}

\begin{table}[!t]
\centering
\setlength\tabcolsep{2.5pt}
\caption{Ablation study on MSConvStar. The schematic diagram of four network structural combinations is shown in Fig.~\ref{figs:msconvstar}}
\label{table:msconvstar}
\scalebox{1}{
\begin{tabular}{ccc|ccc}
\bottomrule
Conv & Star & Multi-Scale & \makecell{Params.\\ (K)} & \makecell{Multi-Adds \\ (G)} & \makecell{Urban100 \\ PSNR (dB) / SSIM} \\ \hline
\xmarkg & \xmarkg & \xmarkg & 710  & 193.6 & 32.98 / 0.9359 \\
\cmark & \xmarkg & \xmarkg & 760 & 204.7 & 33.06 / 0.9367 \\
\cmark & \cmark & \xmarkg & 702 & 191.4 & 33.19 / 0.9379 \\
\cmark & \cmark & \cmark & 694 & 189.6 & \textbf{33.22} / \textbf{0.9381} \\
\toprule
\end{tabular}}
\vspace{-0.3cm}
\end{table}

\begin{table}[!t]
\centering
\setlength\tabcolsep{3pt}
\caption{Ablation study on the multi-range Dependencies. we ensure consistency in model parameters by adjusting the number of channels in the $1\times1$ convolution of MSConvStar.}
\label{table:multi-range}
\scalebox{1}{
\begin{tabular}{ccc|cccc}
\bottomrule
Local & Regional & Sparse Global & \makecell{Multi-Adds \\ (G)} & \makecell{Urban100 \\ PSNR (dB) / SSIM} \\ \hline
\cmark & \cmark & \xmarkg & 189.6 & 32.94 / 0.9358 \\
\cmark & \xmarkg & \cmark & 189.6 &33.00 / 0.9366 \\
\xmarkg & \cmark & \cmark & 189.7 & 33.15 / 0.9376 \\
\cmark & \cmark & \cmark & 189.6 & \textbf{33.22} / \textbf{0.9381}  \\
\toprule
\end{tabular}}
\vspace{-0.3cm}
\end{table}

\begin{table*}[!t]
    \center
    \begin{center}
    \small
     \caption{Quantitative comparison (PSNR (dB) / SSIM) with state-of-the-art methods for \textbf{lightweight SR} on five benchmark datasets. ’Multi-Adds’ is calculated under the setting of upscaling one image to $1280 \times 720$ resolution. For a fair comparison, \textbf{only} the DIV2K dataset is used for training. The best and second best results are marked with bold and underline, respectively. ’N/A’ means that the result is not available.}
    \label{table:benchmark1}
    \setlength{\tabcolsep}{0.9mm} 
	\resizebox{0.94\linewidth}{!}
 {
		\begin{tabular}{|c|l|c|c|c|c|c|c|c|c|} 
			\hline
			Scale & Method  & Annual & \makecell{Params. \\ (K)} & \makecell{Multi-Adds \\ (G)} & \makecell{Set5 \\ PSNR / SSIM} & \makecell{Set14 \\ PSNR / SSIM} & \makecell{B100 \\ PSNR / SSIM} & \makecell{Urban100 \\ PSNR / SSIM} & \makecell{Manga109 \\ PSNR / SSIM} \\[0.5ex] \hline\hline

		\multirow{11}{*}{$\times 2$} & EDSR-baseline~\cite{lim2017enhanced} & CVPRW17 & 1370 & 316.3 & 37.99 / 0.9604 & 33.57 / 0.9175 & 32.16 / 0.8994 & 31.98 / 0.9272 & 38.54 / 0.9769  \\
		& IMDN~\cite{hui2019lightweight} & MM19 & 694 & 158.8 & 38.00 / 0.9605 & 33.63 / 0.9177 & 32.19 / 0.8996 & 32.17 / 0.9283 & 38.88 / 0.9774 \\
		& LatticeNet~\cite{luo2020latticenet} & ECCV20 & 756 & 169.5 & 38.06 / 0.9607 & 33.70 / 0.9187 & 32.20 / 0.8999 & 32.25 / 0.9288 &  N/A \\
            & SwinIR-light~\cite{liang2021swinir} & ICCVW21 & 910 & 252.9 & 38.14 / 0.9611 & 33.86 / 0.9206 & 32.31 / 0.9012 & 32.76 / 0.9340 & 39.12 / 0.9783 \\
            & SwinIR-NG~\cite{choi2023n} & CVPR23 & 1181 & 274.1 & 38.17 / 0.9612 & 33.94 / 0.9205 & 32.31 / 0.9013 & 32.78 / 0.9340 & 39.20 / 0.9781 \\
            & Omni-SR~\cite{wang2023omni} & CVPR23 & 798 &  N/A & 38.22 / 0.9613 & 33.98 / 0.9210 & \underline{32.36} / \underline{0.9020} & \underline{33.05} / \underline{0.9363} & 39.28 / {0.9784} \\
            & SRFormer-light~\cite{zhou2023srformer} & ICCV23 & 853 & 236.2 & \underline{38.23} / 0.9613 & 33.94 / 0.9209 & \underline{32.36} / 0.9019 & 32.91 / 0.9353 & 39.28 / 0.9785 \\
            & MambaIR-light~\cite{guo2024mambair} & ECCV24 & 1363 & 278.9 & 38.16 / 0.9610 & \underline{34.00} / \underline{0.9212} & 32.34 / 0.9017 & 32.92 / 0.9356 & 39.31 / 0.9779 \\
            & \textcolor{yellow}{SRConvNet-L~\cite{li2025srconvnet}} & IJCV25 & 885 & 160 & 38.14 / 0.9610 & 33.81 / 0.9199 & 32.28 / 0.9010 & 32.59 / 0.9321 & 39.22 / 0.9779 \\
            & \textcolor{yellow}{CRAFT~\cite{10852524}} & ICCV23\&PAMI25 & 738 & 197.2 & \underline{38.23} / \underline{0.9615} & 33.92 / 0.9211 & 32.33 / 0.9016 & 32.86 / 0.9343 & \underline{39.39} / \underline{0.9786} \\
            & \textbf{MAT-light} & Ours & \textbf{694} & \textbf{189.6} & \textbf{38.28} / \textbf{0.9617} & \textbf{34.11} / \textbf{0.9227} & \textbf{32.41} / \textbf{0.9029} & \textbf{33.22} / \textbf{0.9381} & \textbf{39.46} / \textbf{0.9789} \\ \hline\hline

            \multirow{11}{*}{$\times 3$} & EDSR-baseline~\cite{lim2017enhanced} & CVPRW17 & 1555 & 160.2 & 34.37 / 0.9270 & 30.28 / 0.8417 & 29.09 / 0.8052 & 28.15 / 0.8527 & 33.45 / 0.9439 \\
		& IMDN~\cite{hui2019lightweight} & MM19 & 703 & 71.5 & 34.36 / 0.9270 & 30.32 / 0.8417 & 29.09 / 0.8046 & 28.17 / 0.8519 & 33.61 / 0.9445 \\
            & LatticeNet~\cite{luo2020latticenet} & ECCV20 & 765 & 76.3 & 34.40 / 0.9272 & 30.32 / 0.8416 & 29.10 / 0.8049 & 28.19 / 0.8513 &  N/A \\
            & SwinIR-light~\cite{liang2021swinir} & ICCVW21 & 918 & 114.5 & 34.62 / 0.9289 & 30.54 / 0.8463 & 29.20 / 0.8082 & 28.66 / 0.8624 & 33.98 / 0.9478 \\
            & SwinIR-NG~\cite{choi2023n} & CVPR23 & 1190 & 114.1 & 34.64 / 0.9293 & 30.58 / 0.8471 & 29.24 / 0.8090 & 28.75 / 0.8639 & 34.22 / 0.9488 \\
            & Omni-SR~\cite{wang2023omni} & CVPR23 & 807 &  N/A & 34.70 / 0.9294 & 30.57 / 0.8469 & 29.28 / 0.8094 & 28.84 / 0.8656 & 34.22 / 0.9487 \\
            & SRFormer-light~\cite{zhou2023srformer} & ICCV23 & 861 & 104.8 & 34.67 / \underline{0.9296} & 30.57 / 0.8469 & 29.26 / 0.8099 & 28.81 / 0.8655 & 34.19 / 0.9489 \\
            & MambaIR-light~\cite{guo2024mambair} & ECCV24 & 1371 & 124.6 & \underline{34.72} / \underline{0.9296} & \underline{30.63} / \underline{0.8475} & \underline{29.29} / \underline{0.8099} & \underline{29.00} / \underline{0.8689} & \underline{34.39} / \underline{0.9495} \\
            & \textcolor{yellow}{SRConvNet-L~\cite{li2025srconvnet}} & IJCV25 & 906 & 74 & 34.59 / 0.9288 & 30.50 / 0.8455 & 29.22 / 0.8081 & 28.56 / 0.8600 & 34.17 / 0.9479 \\
            & \textcolor{yellow}{CRAFT~\cite{10852524}} & ICCV23\&PAMI25 & 744 & 87.5 & 34.71 / 0.9295 & 30.61 / 0.8469 & 29.24 / 0.8093 & 28.77 / 0.8635 & 34.29 / 0.9491 \\
            & \textbf{MAT-light} & Ours & \textbf{703} & \textbf{85.0} & \textbf{34.79} / \textbf{0.9303} & \textbf{30.68} / \textbf{0.8491} & \textbf{29.32} / \textbf{0.8116} & \textbf{29.03} / \textbf{0.8698} & \textbf{34.49} / \textbf{0.9505}\\ \hline\hline

            \multirow{11}{*}{$\times 4$} & EDSR-baseline~\cite{lim2017enhanced} & CVPRW17 & 1518 & 114.0 & 32.09 / 0.8938 & 28.58 / 0.7813 & 27.57 / 0.7357 & 26.04 / 0.7849 & 30.35 / 0.9067  \\
		& IMDN~\cite{hui2019lightweight} & MM19 & 715 & 40.9 & 32.21 / 0.8948 & 28.58 / 0.7811 & 27.56 / 0.7353 & 26.04 / 0.7838 & 30.45 / 0.9075 \\
		& LatticeNet~\cite{luo2020latticenet} & ECCV20 & 777 & 43.6 & 32.18 / 0.8943 & 28.61 / 0.7812 & 27.57 / 0.7355 & 26.14 / 0.7844 &  N/A \\
            & SwinIR-light~\cite{liang2021swinir} & ICCVW21 & 930 & 65.2 & 32.44 / 0.8976 & 28.77 / 0.7858 & 27.69 / 0.7406 & 26.47 / 0.7980 & 30.92 / 0.9151 \\
            & SwinIR-NG~\cite{choi2023n} & CVPR23 & 1201 & 63.0 & 32.44 / 0.8980 & 28.83 / 0.7870 & 27.73 / 0.7418 & 26.61 / 0.8010 & 31.09 / 0.9161 \\
            & Omni-SR~\cite{wang2023omni} & CVPR23 & 819 & N/A & 32.49 / 0.8988 & 28.78 / 0.7859 & 27.71 / 0.7415 & 26.64 / 0.8018 & 31.02 / 0.9151 \\
            & SRFormer-light~\cite{zhou2023srformer} & ICCV23 & 873 & 62.8 & 32.51 / 0.8988 & 28.82 / 0.7872 & 27.73 / 0.7422 & 26.67 / 0.8032 & 31.17 / 0.9165 \\
            & MambaIR-light~\cite{guo2024mambair} & ECCV24 & 1383 & 70.8 & 32.51 / \underline{0.8993} & \underline{28.85} / \underline{0.7876} & \underline{27.75} / \underline{0.7423} & \underline{26.75} / \underline{0.8051} & \underline{31.26} / \underline{0.9175} \\
            & \textcolor{yellow}{SRConvNet-L~\cite{li2025srconvnet}}  & IJCV25 & 902 & 45 & 32.44 / 0.8976 & 28.77 / 0.7857 & 27.69 / 0.7402 & 26.47 / 0.7970 & 30.96 / 0.9139 \\
            & \textcolor{yellow}{CRAFT~\cite{10852524}} & ICCV23\&PAMI25 & 753 & 52.4 & \underline{32.52} / 0.8989 & \underline{28.85} / 0.7872 & 27.72 / 0.7418 & 26.56 / 0.7995 & 31.18 / 0.9168 \\
            & \textbf{MAT-light} & Ours & \textbf{714}  & \textbf{48.5} & \textbf{32.61} / \textbf{0.8998} & \textbf{28.92} / \textbf{0.7897} & \textbf{27.79} / \textbf{0.7444} & \textbf{26.83} / \textbf{0.8088} & \textbf{31.38} / \textbf{0.9192} \\

		\hline
	\end{tabular}}
    \end{center}
\end{table*}

\begin{figure}[t!]
\centering
\setlength{\abovecaptionskip}{0.2cm}
\includegraphics[width=0.99\linewidth]{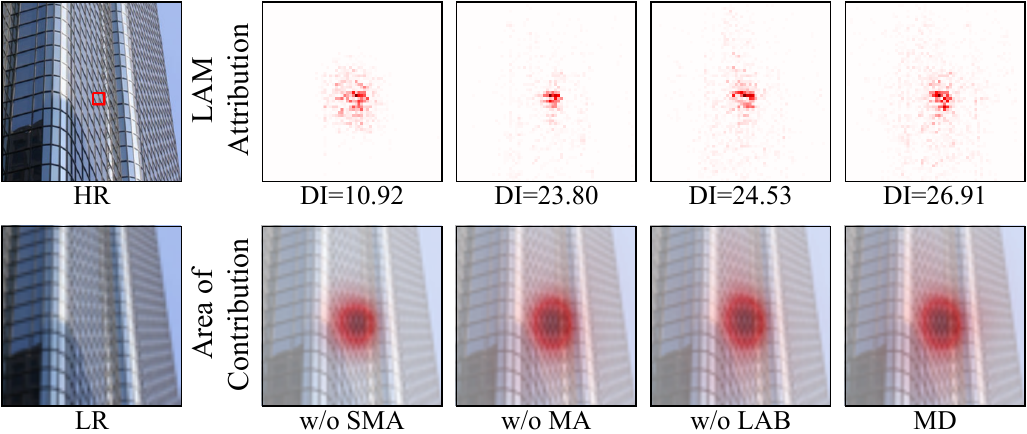}
\caption{\textcolor{yellow}{LAM~\cite{gu2021interpreting} results under different range dependency configurations.}}
\label{figs:LAM_mrf}
\vspace{-0.3cm}
\end{figure}

\begin{figure}[!ht]
\centering
\includegraphics[width=0.99\linewidth]{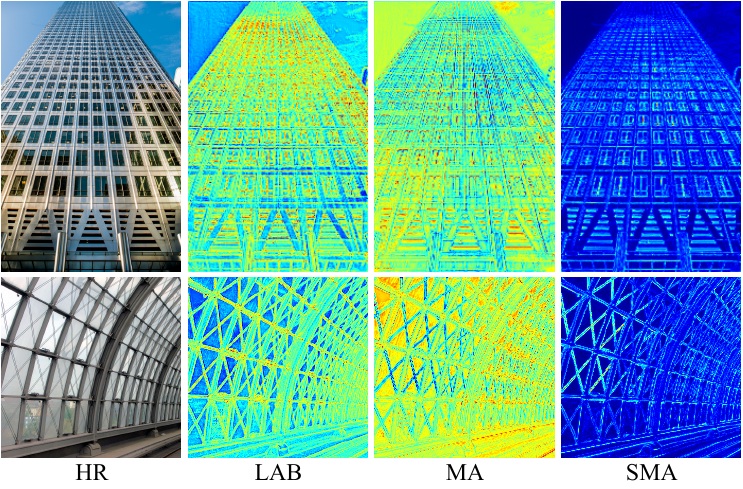}
\caption{\textcolor{yellow}{Visualization of feature maps for multi-range dependencies. LAB and MA emphasize more on low-frequency structural information, while SMA focuses more on high-frequency edge details.}}
\label{figs:feature2}
\vspace{-0.3cm}
\end{figure}

\noindent \textbf{Effectiveness of Multi-Range Dependencies.}
MAT integrates multi-range features through three specialized components: local convolutions for local features, MA for regional features, and SMA for sparse global features. We evaluate four different feature integration schemes (Table~\ref{table:multi-range}), maintaining consistent model parameters by adjusting MSConvStar's $1\times1$ convolution channels. Experimental results demonstrate that omitting any feature hierarchy level degrades model performance, with larger-range features showing particularly significant impact.

\textcolor{yellow}{We further analyze component contributions through LAM visualization across four model configurations (Fig.~\ref{figs:LAM_mrf}). Results indicate that removing any feature extraction mechanism diminishes model perceptual capabilities, with SMA removal causing the most significant reduction in perception range, while MA or LAB removal yields more modest decrements. This pattern suggests SMA's primary role in non-local information processing, contrasting with LAB and MA's focus on local feature extraction. Feature map analysis (Fig.~\ref{figs:feature2}) reveals complementary roles: LAB and MA mainly capture low-frequency structural information, while SMA specializes in high-frequency edge details, enabling MAT to effectively model multi-range dependencies through synergistic feature extraction.}

\begin{figure*}[!t]
\centering
\setlength{\abovecaptionskip}{0cm}
\includegraphics[width=0.94\textwidth]{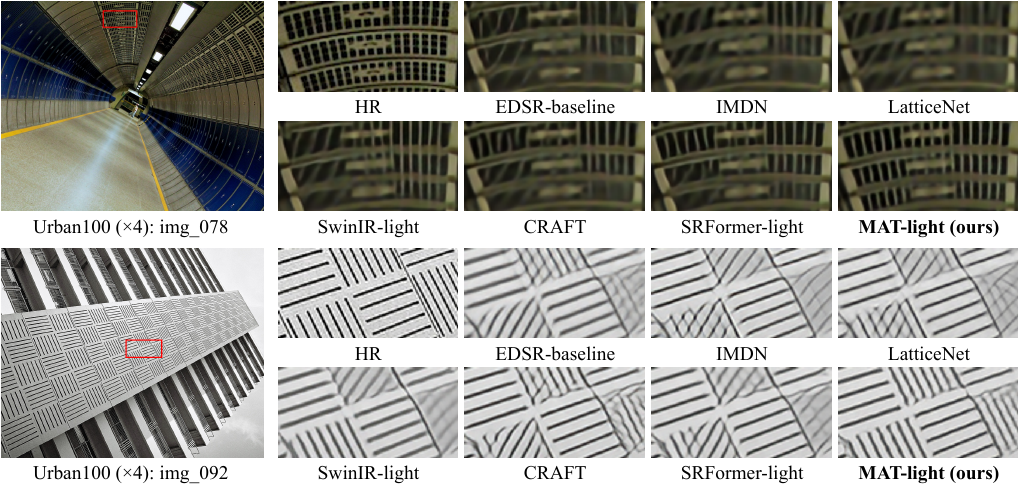}
\caption{Visual comparison on $\times 4$ \textbf{lightweight image SR}. The patches for comparison are marked with red boxes in the original images.}
\label{figs:visual2}
\vspace{-0.4cm}
\end{figure*}

\begin{table*}[!t]
    \center
    \small
    \begin{center}
    \caption{Quantitative comparison (PSNR (dB) / SSIM) with state-of-the-art methods for \textbf{classical SR} on five benchmark datasets. ’Multi-Adds’ is calculated under the setting of upscaling one image to $256^2$ resolution. The best and second best results are marked in bold and underline, respectively. ’N/A’ means that the result is not available.}
	\label{table:benchmark2}
	\setlength{\tabcolsep}{0.9mm}
    \resizebox{0.94\linewidth}{!}
	{
		\begin{tabular}{|c|l|c|c|c|c|c|c|c|c|} 
			\hline
			Scale & Method & Annual & \makecell{Params. \\ (M)} & \makecell{Multi-Adds \\ (G)} & \makecell{Set5 \\ PSNR / SSIM} & \makecell{Set14 \\ PSNR / SSIM} & \makecell{B100 \\ PSNR / SSIM} & \makecell{Urban100 \\ PSNR / SSIM} & \makecell{Manga109 \\ PSNR / SSIM} \\[0.5ex] \hline\hline

		\multirow{13}{*}{$\times 2$} & EDSR~\cite{lim2017enhanced} & CVPRW17  & 40.73 & 667.4 & 38.11 / 0.9602 & 33.92 / 0.9195 & 32.32 / 0.9013 & 32.93 / 0.9351 & 39.10 / 0.9773 \\
            & RCAN~\cite{zhang2018image} & ECCV18 & 15.44 & 251.0 & 38.27 / 0.9614 & 34.12 / 0.9216 & 32.41 / 0.9027 & 33.34 / 0.9384 & 39.44 / 0.9786 \\
		& IGNN~\cite{zhou2020cross} & NeurIPS20 & 49.51 & N/A & 38.24 / 0.9613 & 34.07 / 0.9217 & 32.41 / 0.9025 & 33.23 / 0.9383 & 39.35 / 0.9786 \\
		& NLSN~\cite{mei2021image} & CVPR21 & 41.80 & 731.4 & 38.34 / 0.9618 & 34.08 / 0.9231 & 32.43 / 0.9027 & 33.42 / 0.9394 & 39.59 / 0.9789 \\
            & SwinIR~\cite{liang2021swinir} & ICCVW21 & 11.75 & 205.3 & 38.42 / 0.9623 & 34.46 / 0.9250 & 32.53 / 0.9041 & 33.81 / 0.9427 & 39.92 / 0.9797 \\
            & ART-S~\cite{zhang2023accurate} & ICLR23 & 11.71 & 227.6 & 38.48 / 0.9625 & 34.50 / 0.9258 & 32.53 / 0.9043 & 34.02 / 0.9437 & 40.11 / 0.9804 \\
            & \textcolor{yellow}{HAT-S~\cite{chen2023activating}} & CVPR23 & 9.47 & 209.6 & 38.58 / 0.9628 & \textbf{34.70} / 0.9261 & 32.59 / 0.9050 & \underline{34.31} / 0.9459 & 40.14 / 0.9805 \\
            & \textcolor{yellow}{DAT-S~\cite{chen2023dual}} & ICCV23  & 11.06 & 193.3 & 38.54 / 0.9627 & 34.60 / 0.9258 & 32.57 / 0.9047 & 34.12 / 0.9444 & 40.17 / 0.9804 \\
            & SRFormer~\cite{zhou2023srformer} & ICCV23 & 10.38 & 206.4 & 38.51 / 0.9627 & 34.44 / 0.9253 & 32.57 / 0.9046 & 34.09 / 0.9449 & 40.07 / 0.9802 \\
            & \textcolor{yellow}{FDRNet~\cite{10643469}} & TIP24 & 8.66 & N/A & 38.54 / 0.9627 & 34.65 / \underline{0.9262} & 32.57 / 0.9046 & 34.18 / 0.9449 & 40.10 / 0.9803 \\
            & MambaIR~\cite{guo2024mambair} & ECCV24 & 20.42 & 318.8 & 38.57 / 0.9627 & \underline{34.67} / 0.9261 & 32.58 / 0.9048 & 34.15 / 0.9446 & \underline{40.28} / 0.9806 \\
            & \textbf{MAT} & Ours & 9.60 & 187.0 & \underline{38.61} / \underline{0.9631} & 34.53 / 0.9259 & \underline{32.62} / \underline{0.9053} & \underline{34.31} / \underline{0.9462} & 40.22 / \underline{0.9808} \\ 
            & \textbf{MAT+} & Ours & 9.60 & 187.0 & \textbf{38.65} / \textbf{0.9632} & 34.60 / \textbf{0.9263} & \textbf{32.65} / \textbf{0.9056} & \textbf{34.44} / \textbf{0.9468} & \textbf{40.33} / \textbf{0.9811} \\ \hline\hline

            \multirow{13}{*}{$\times 3$} & EDSR~\cite{lim2017enhanced} & CVPRW17 & 43.68 & 315.9 & 34.65 / 0.9280 & 30.52 / 0.8462 & 29.25 / 0.8093 & 28.80 / 0.8653 & 34.17 / 0.9476 \\
            & RCAN~\cite{zhang2018image} & ECCV18 & 15.63 & 112.1 & 34.74 / 0.9299 & 30.65 / 0.8482 & 29.32 / 0.8111 & 29.09 / 0.8702 & 34.44 / 0.9499 \\
		& IGNN~\cite{zhou2020cross} & NeurIPS20 & 49.51 & N/A & 34.72 / 0.9298 & 30.66 / 0.8484 & 29.31 / 0.8105 & 29.03 / 0.8696 & 34.39 / 0.9496 \\
		& NLSN~\cite{mei2021image} & CVPR21 & 44.75 & 344.1 & 34.85 / 0.9306 & 30.70 / 0.8485 & 29.34 / 0.8117 & 29.25 / 0.8726 & 34.57 / 0.9508 \\
            & SwinIR~\cite{liang2021swinir} & ICCV21 & 11.94 & 98.5 & 34.97 / 0.9318 & 30.93 / 0.8534 & 29.46 / 0.8145 & 29.75 / 0.8826 & 35.12 / 0.9537 \\
            & ART-S~\cite{zhang2023accurate} & ICLR23 & 11.90 & 102.0 & 34.98 / 0.9318 & 30.94 / 0.8530 & 29.45 /0.8146 & 29.86 / 0.8830 & 35.22 / 0.9539 \\
            & \textcolor{yellow}{HAT-S~\cite{chen2023activating}} & CVPR23 & 9.66 & 119.7 & 35.01 / 0.9325 & \underline{31.05} / \underline{0.8550} & 29.50 / 0.8158 & \underline{30.15} / \underline{0.8879} & 35.40 / 0.9547 \\
            & \textcolor{yellow}{DAT-S~\cite{chen2023dual}} & ICCV23  & 11.25 & 88.3 & \underline{35.12} / 0.9327 & 31.04 / 0.8543 & 29.51 / 0.8157 & 29.98 / 0.8846 & 35.41 / 0.9546 \\
            & SRFormer~\cite{zhou2023srformer} & ICCV23 & 10.56 & 93.2 & 35.02 / 0.9323 & 30.94 / 0.8540 & 29.48 / 0.8156 & 30.04 / 0.8865 & 35.26 / 0.9543 \\
            & \textcolor{yellow}{FDRNet~\cite{10643469}} & TIP24 & 8.85 & N/A & 34.98 / 0.9320 & 30.98 / 0.8535 & 29.47 / 0.8152 & 29.97 / 0.8856 & 35.24 / 0.9540 \\
            & MambaIR~\cite{guo2024mambair} & ECCV24 & 20.61 & 142.0 & 35.08 / 0.9323 & 30.99 / 0.8536 & 29.51 / 0.8157 & 29.93 / 0.8841 & \underline{35.43} / 0.9546 \\
            & \textbf{MAT} & Ours & 9.78 & 83.9 & 35.09 / \underline{0.9328} & 31.03 / \underline{0.8550} & \underline{29.53} / \underline{0.8167} & 30.11 / \underline{0.8879} & 35.41 / \underline{0.9549} \\
            & \textbf{MAT+} & Ours & 9.78 & 83.9 & \textbf{35.13} / \textbf{0.9330} & \textbf{31.08} / \textbf{0.8555} & \textbf{29.56} / \textbf{0.8171} & \textbf{30.22} / \textbf{0.8893} & \textbf{35.55} / \textbf{0.9554} \\ \hline\hline

		\multirow{13}{*}{$\times 4$} & EDSR~\cite{lim2017enhanced} & CVPRW17 & 43.09 & 205.8 & 32.46 / 0.8968 & 28.80 / 0.7876 & 27.71 / 0.7420 & 26.64 / 0.8033 & 31.02 / 0.9148 \\
            & RCAN~\cite{zhang2018image} & ECCV18 & 15.59 & 65.3 & 32.63 / 0.9002 & 28.87 / 0.7889 & 27.77 / 0.7436 & 26.82 / 0.8087 & 31.22 / 0.9173 \\
		& IGNN~\cite{zhou2020cross} & NeurIPS20 & 49.51 & N/A & 32.57 / 0.8998 & 28.85 / 0.7891 & 27.77 / 0.7434 & 26.84 / 0.8090 & 31.28 / 0.9182 \\
		& NLSN~\cite{mei2021image} & CVPR21 & 44.16 & 221.8 & 32.59 / 0.9000 & 28.87 / 0.7891 & 27.78 / 0.7444 & 26.96 / 0.8109 & 31.27 / 0.9184 \\
            & SwinIR~\cite{liang2021swinir} & ICCVW21 & 11.90 & 53.8 & 32.92 / 0.9044 & 29.09 / 0.7950 & 27.92 / 0.7489 & 27.45 / 0.8254 & 32.03 / 0.9260 \\
            & ART-S~\cite{zhang2023accurate} & ICLR23 & 11.87 & 55.6 & 32.86 / 0.9029 & 29.09 / 0.7942 & 27.91 / 0.7489 & 27.54 / 0.8261 & 32.13 / 0.9263 \\
            & \textcolor{yellow}{HAT-S~\cite{chen2023activating}} & CVPR23 & 9.61 & 54.9 & 32.92 / 0.9047 & 29.15 / 0.7958 & 27.97 / 0.7505 & \underline{27.87} / \underline{0.8346} & \underline{32.35} / \underline{0.9283} \\
            & \textcolor{yellow}{DAT-S~\cite{chen2023dual}} & ICCV23  & 11.21 & 50.8 & 33.00 / 0.9047 & \underline{29.20} / \underline{0.7962} & 27.97 / 0.7510 & 27.68 / 0.8300 & 32.33 / 0.9278 \\
            & SRFormer~\cite{zhou2023srformer} & ICCV23 & 10.52 & 54.3 & 32.93 / 0.9041 & 29.08 / 0.7953 & 27.94 / 0.7502 & 27.68 / 0.8311 & 32.21 / 0.9271 \\
            & \textcolor{yellow}{FDRNet~\cite{10643469}} & TIP24 & 8.81 & N/A & 32.82 / 0.9339 & 29.11 / 0.7947 & 27.94 / 0.7494 & 27.63 / 0.8301 & 32.17 / 0.9264 \\
            & MambaIR~\cite{guo2024mambair} & ECCV24 & 20.57 & 82.2 & 33.03 / 0.9046 & \underline{29.20} / 0.7961 & 27.98 / 0.7503 & 27.68 / 0.8287 & 33.32 / 0.9272 \\
            & \textbf{MAT} & Ours & 9.74 & 49.2 & \underline{33.06} / \underline{0.9054} & 29.17 / 0.7960 & \underline{27.99} / \underline{0.7514} & 27.78 / 0.8328 & 32.31 / 0.9282 \\
            & \textbf{MAT+} & Ours & 9.74 & 49.2 & \textbf{33.08} / \textbf{0.9056} & \textbf{29.24} / \textbf{0.7972} & \textbf{28.02} / \textbf{0.7520} & \textbf{27.89} / \textbf{0.8349} & \textbf{32.49} / \textbf{0.9294} \\

		\hline
	\end{tabular}}
    \end{center}
\end{table*}

\begin{figure*}[!t]
\centering
\includegraphics[width=0.9\textwidth]{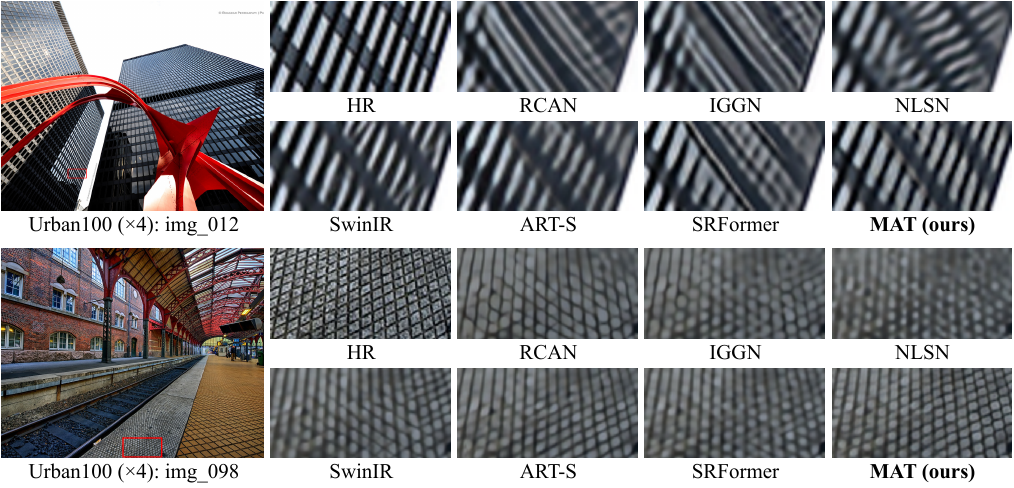}
\caption{Visual comparison on $\times 4$ \textbf{classical image SR}. The patches for comparison are marked with red boxes in the original images.}
\label{figs:visual3}
\vspace{-0.2cm}
\end{figure*}

\subsection{Lightweight Image Super-Resolution}

\textcolor{yellow}{To demonstrate the effectiveness and efficiency of proposed MAT, we compare our MAT-light with various recent state-of-the-art lightweight SR methods, including the EDSR-baseline~\cite{lim2017enhanced}, IMDN~\cite{hui2019lightweight}, LatticeNet~\cite{luo2020latticenet}, SwinIR-light~\cite{liang2021swinir}, SwinIR-NG~\cite{choi2023n}, Omni-SR~\cite{wang2023omni}, SRFormer-light~\cite{zhou2023srformer}, MambaIR-light~\cite{guo2024mambair}, SRConvNet-L~\cite{li2025srconvnet} and CRAFT~\cite{10852524}.}

\noindent \textbf{Quantitative Comparison.}
Table~\ref{table:benchmark1} presents a quantitative comparison of lightweight image SR models. We report both model parameters and Multi-Adds to measure computational efficiency and complexity. Our MAT-light demonstrates the best performance across all five benchmark datasets for various scale factors while maintaining the lowest parameter count and competitive computational complexity. Notably, compared to CNN-based approaches, MAT-light achieves substantial improvements of approximately \textbf{0.8dB} and \textbf{0.9dB} on Urban100 and Manga109 datasets, respectively. When compared to Transformer-based models, MAT-light shows significant gains of \textbf{0.2dB$\sim$0.46dB} on the Manga109 dataset. Furthermore, our model outperforms the recent Mamba-based architecture, MambaIR-light, consistently across all evaluation metrics.

\noindent \textbf{Visual Comparison.}
As shown in Fig.~\ref{figs:visual1} and~\ref{figs:visual2}, we compare the visual results between MAT-light and other lightweight image SR models. MAT-light successfully recovers fine lines and texture details in the images, while others produce blurry artifacts and inaccurate details. These visual results demonstrate MAT-light's superior reconstruction capability through its effective feature extraction across different spatial ranges.

\subsection{Classical Image Super-Resolution}

\textcolor{yellow}{To further demonstrate the scalability of MAT, we expand MAT to build a large model and compare it with a series of state-of-the-art classical SR methods: EDSR~\cite{lim2017enhanced}, RCAN~\cite{zhang2018image}, IGNN~\cite{zhou2020cross}, NLSN~\cite{mei2021image}, SwinIR~\cite{liang2021swinir}, ART~\cite{zhang2023accurate}, HAT~\cite{chen2023activating}, DAT~\cite{chen2023dual}, SRFormer~\cite{zhou2023srformer}, FDRNet~\cite{10643469} and MambaIR~\cite{guo2024mambair}. Consistent with prior works~\cite{liang2021swinir, zhang2023accurate, zhou2023srformer}, self-ensemble strategy is introduced in testing to further improve the model's performance, denoted by the symbol “+”.}

\noindent \textbf{Quantitative Comparison.}
Table~\ref{table:benchmark2} shows the quantitative comparison of classical image SR models. MAT achieves superior performance across all five benchmark datasets and scale factors while requiring minimal parameters and computational complexity. Through its multi-range attention mechanism and feature modeling, MAT effectively captures hierarchical image features. The performance gains are further enhanced when applying the self-ensemble strategy, demonstrating MAT's effectiveness and scalability.

\noindent \textbf{Visual Comparison.}
Fig.~\ref{figs:visual3} presents the visual comparison between MAT and other classical image SR models. MAT accurately reconstructs the main image structures with fewer blur artifacts compared to other methods. This superior reconstruction quality stems from MAT's ability to capture features at multiple spatial ranges.

\begin{table*}[!t]
\renewcommand{\arraystretch}{1}
\begin{center}
\caption{The model depth and performance comparisons on $\times 4$ SR. ’Multi-Adds’, 'Running Time' and 'Memory' is calculated under the setting of upscaling one image to $1280 \times 720$ resolution. The best and second best results are marked in \textcolor{red}{red} and \textcolor{blue}{blue} colors, respectively.}
\vspace{-0.2cm}
\setlength\tabcolsep{2pt}
\scalebox{1}{
    \begin{tabular}{|l|c|c|c|c|c|c|c|c|c|c|}
        \hline
         Method & \makecell{\#Params. \\ (K)} & \makecell{Multi-Adds \\ (G)}  & \textbf{\makecell{ \#Depth \\ Layers (Blocks)}} & \makecell{\#Time \\ (ms)}& \makecell{\#Mem. \\ (G)} & \makecell{Set5 \\ PSNR / SSIM} & \makecell{Set14 \\ PSNR / SSIM} & \makecell{B100 \\ PSNR / SSIM} & \makecell{Urban100 \\ PSNR / SSIM} & \makecell{Manga109 \\ PSNR / SSIM} \\ \hline\hline
         SwinIR-light & 930 & 65.2  & \textbf{103 (24)} & 143.5 & \textcolor{blue}{6.3} & 32.44 / 0.8980 & 28.83 / 0.7870 & 27.73 / 0.7418 & 26.47 / 0.7980 & 30.92 / 0.9151 \\ 
        SRFormer-light & 873 & 62.8 & \textbf{127 (24)} & 162.9 & 7.4 & 32.51 / 0.8988 & 28.82 / 0.7872 & 27.73 / 0.7422 & 26.67 / 0.8032 & 31.17 / 0.9165 \\
        \textbf{MAT-light} & 714 & 48.5 & \textbf{107 (20)} & \textcolor{red}{71.1} & \textcolor{red}{5.3} & \textcolor{blue}{32.61} / \textcolor{blue}{0.8998} & \textcolor{blue}{28.92} / \textcolor{blue}{0.7897} & \textcolor{blue}{27.79} / \textcolor{blue}{0.7444} & \textcolor{blue}{26.83} / \textcolor{blue}{0.8088} & \textcolor{blue}{31.38} / \textcolor{blue}{0.9192} \\
        \textbf{MAT-light-V2} & 956 & 66.4 & \textbf{147 (28)} & \textcolor{blue}{105.5} & 7.6 & \textcolor{red}{32.68} / \textcolor{red}{0.9010} & \textcolor{red}{29.01} / \textcolor{red}{0.7915} & \textcolor{red}{27.84} / \textcolor{red}{0.7460} & \textcolor{red}{26.99} / \textcolor{red}{0.8129} & \textcolor{red}{31.51} / \textcolor{red}{0.9206}   \\
        \hline
    \end{tabular}}
\label{table:depth}
\end{center}
\end{table*}

\begin{figure*}[!t]
\centering
\includegraphics[width=0.98\textwidth]{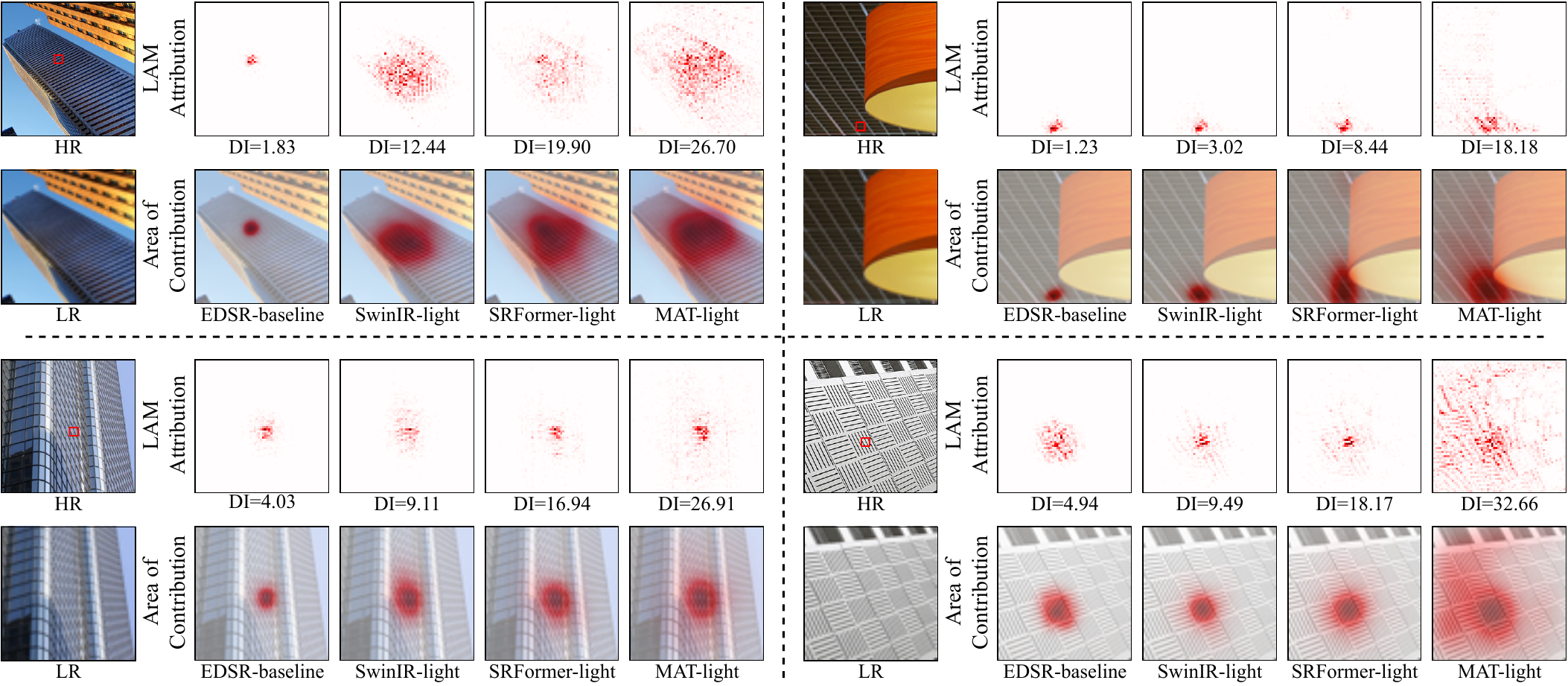}
\caption{Comparison of LAM results of EDSR-baseline~\cite{lim2017enhanced}, SwinIR-light~\cite{liang2021swinir}, SRFormer-light~\cite{zhou2023srformer}, and the proposed MAT-light.}
\label{figs:lam_more}
\vspace{-0.4cm}
\end{figure*}

\begin{figure}[!t]
\centering
\includegraphics[width=0.98\linewidth]{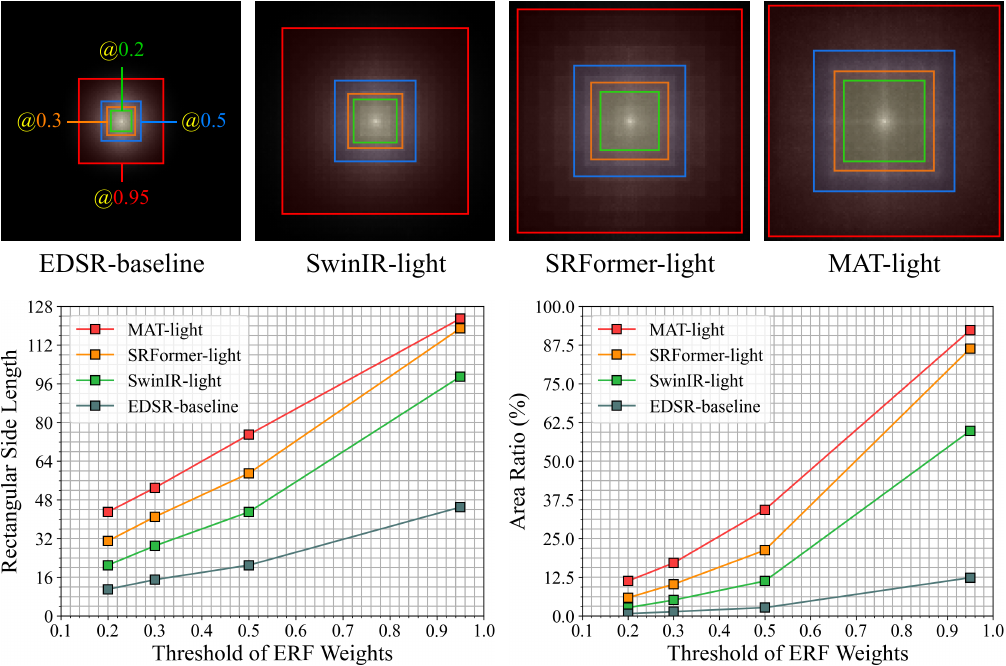}
\caption{The efective receptive field (ERF)~\cite{ding2022scaling} visualization and comparison for EDSR-baseline~\cite{lim2017enhanced}, SwinIR-light~\cite{liang2021swinir}, SRFormer-light~\cite{zhou2023srformer}, and the proposed MAT-light. A larger ERF is indicated by a more extensively distributed bright area.}
\label{figs:erf}
\vspace{-0.2cm}
\end{figure}

\subsection{Model Depth Analyses}
The depth of the model exerts a significant influence on computational efficiency. Table~\ref{table:depth} presents a comparative analysis of model depth and performance. MAT achieves competitive performance and efficiency in its current depth configuration through the multi-range attention mechanism. Notably, by increasing the number of RMAGs to 6 in MAT-light to align its capacity with comparable architectures, our MAT-light-V2 achieves greater performance gains while maintaining a shorter runtime. These results demonstrate the computational efficiency of our proposed MAT architecture.

\subsection{Analyses of LAM and ERF}
We use local attribution maps (LAM)~\cite{gu2021interpreting} to visualize the spatial range of information used in target area reconstruction. \textcolor{yellow}{As shown in Fig.~\ref{figs:lam_more}, MAT-light effectively identifies and focuses on structurally similar regions across non-local spatial ranges while suppressing dissimilar features, whereas other methods are limited to local feature extraction.} Quantitatively, MAT-light achieves significantly higher diffusion index (DI) values compared to other models. We also evaluate the effective receptive field (ERF)~\cite{ding2022scaling} to measure the actual receptive field. Fig.~\ref{figs:erf} shows the comparison of ERF rectangular side length and area ratio under different weight thresholds. MAT-light demonstrates a larger ERF, enabling broader information perception. These results validate the effectiveness of the multi-range representation learning.

\subsection{Running Time and Memory Comparisons}
\begin{figure}[!t]
\centering
\includegraphics[width=0.98\linewidth]{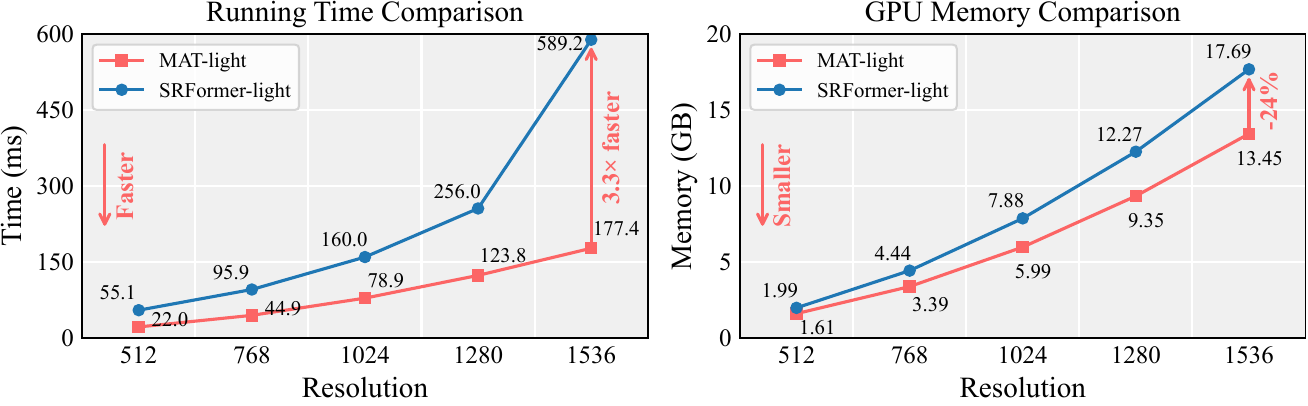}
\caption{Running time and memory comparisons on $\times 4$ SR.}
\label{figs:speed_memory}
\vspace{-0.2cm}
\end{figure}

We evaluate the computational efficiency of MAT-light with the representative Transformer-based model SRFormer-light~\cite{zhou2023srformer} for $\times 4$ SR tasks. We measure average running time across 100 test images and peak GPU memory consumption during inference on an RTX 3090. As shown in Fig.~\ref{figs:speed_memory}, SRFormer-light requires longer running time and higher memory usage due to its large window sizes, which involve additional padding pixels and shift operations. In contrast, MAT-light demonstrates significant efficiency gains. For generating an HR image of $1536^2$ resolution, MAT-light runs approximately $3.3\times$ faster than SRFormer-light while reducing GPU memory consumption by $24\%$.

\section{Conclusion}
In this paper, we present MAT, a highly effective model for image SR. MAT combines dilated operations with self-attention mechanisms to implement multi-range attention, enabling flexible attention scopes and enhanced perception of both regional and sparse global features. By integrating this with local feature extraction, MAT achieves effective multi-range representation learning. We also introduce the MSConvStar module, a streamlined enhancement to image token interconnections. Thorough experiments validate MAT's effectiveness and efficiency, with MAT achieving state-of-the-art performance in both lightweight and classical SR while requiring fewer parameters and lower computational costs.

\bibliographystyle{IEEEtran}

\bibliography{IEEEabrv, main}

\end{document}